\documentclass[%
 aip,
 cp,  
 amsmath,amssymb,
 reprint,%
]{revtex4-2}

\usepackage{graphicx}
\usepackage{dcolumn}
\usepackage{bm}
\usepackage[utf8]{inputenc}
\usepackage[T1]{fontenc}
\usepackage{mathptmx} 
\usepackage[bottom]{footmisc}

\usepackage{url}
\usepackage{hyperref}
\hypersetup{colorlinks=true,linkcolor=blue,citecolor=blue,urlcolor=blue}

\usepackage{graphicx}  
\usepackage{amsmath}   

\begin{document}

\title{A Multi-Resolution Benchmark Framework for Spatial Reasoning Assessment in Neural Networks}

\author{Manuela Imbriani}
\email{manuela.imbriani@isti.cnr.it}
\affiliation{Dipartimento di Fisica, Università di Pisa, Pisa, ITALY}
\affiliation{Istituto di Scienza e Tecnologie dell'Informazione "A.Faedo", CNR, Pisa, ITALY}

\author{Gina Belmonte}
\affiliation{Azienda Toscana Nord Ovest, S.C.Fisica Sanitaria Nord, Lucca, ITALY}

\author{Mieke Massink}
\affiliation{Istituto di Scienza e Tecnologie dell'Informazione "A.Faedo", CNR, Pisa, ITALY}

\author{Alessandro Tofani}
\affiliation{Azienda Toscana Nord Ovest, S.C.Fisica Sanitaria Nord, Lucca, ITALY}

\author{Vincenzo Ciancia}
\affiliation{Istituto di Scienza e Tecnologie dell'Informazione "A.Faedo", CNR, Pisa, ITALY}

\begin{abstract}
This paper presents preliminary results in the definition of a comprehensive benchmark framework designed to systematically evaluate spatial reasoning capabilities in neural networks, with a particular focus on morphological properties such as connectivity and distance relationships.\\
The framework is currently being used to study the capabilities of nnU-Net, exploiting the spatial model checker VoxLogicA to generate two distinct categories of synthetic datasets: maze connectivity problems for topological analysis and spatial distance computation tasks for geometric understanding. Each category is evaluated across multiple resolutions to assess scalability and generalization properties. The automated pipeline encompasses a complete machine learning workflow including: synthetic dataset generation, standardized training with cross-validation, inference execution, and comprehensive evaluation using Dice coefficient and IoU (Intersection over Union) metrics. Preliminary experimental results demonstrate significant challenges in neural network spatial reasoning capabilities, revealing systematic failures in basic geometric and topological understanding tasks. The framework provides a reproducible experimental protocol, enabling researchers to identify specific limitations. Such limitations could be addressed through hybrid approaches combining neural networks with symbolic reasoning methods for improved spatial understanding in clinical applications, establishing a foundation for ongoing research into neural network spatial reasoning limitations and potential solutions.
\end{abstract}

\maketitle

\section{Introduction}
Spatial reasoning represents a fundamental capability that enables understanding of geometric relationships, topological properties, and spatial connectivity patterns. In the context of artificial intelligence and neural networks, developing robust spatial reasoning capabilities is crucial for applications ranging from robotics and autonomous navigation to medical image analysis and computer vision. However, the evaluation of spatial reasoning abilities in neural networks is limited by the lack of standardized benchmark frameworks that can systematically assess these capabilities across different scales and complexity levels. 
Existing evaluation approaches primarily focus on improving network architectures or loss functions, but lack systematic frameworks that can isolate and test specific spatial reasoning capabilities using mathematically precise ground truth.
Recent work has demonstrated fundamental limitations in current neural architectures for spatial reasoning tasks, highlighting the need for specialized evaluation frameworks that can systematically identify these deficiencies.
This paper presents preliminary results from an ongoing effort to develop a comprehensive multi-task benchmark framework designed to systematically evaluate spatial reasoning capabilities in neural networks across a wide spectrum of morphological properties. The complete benchmark framework will encompass numerous test categories addressing various spatial characteristics including the shape and size of connected regions, connectivity patterns, distance relationships, spatial inclusion properties, region counting, and other fundamental geometric and topological features. The current work focuses on presenting initial findings from two representative task categories—maze connectivity analysis and spatial distance computation—as foundational components of this broader evaluation framework.

The reliability of neural networks in medical imaging tasks has become increasingly important as these systems are deployed in clinical environments. While neural networks excel at pattern recognition and feature extraction, their ability to understand fundamental spatial relationships such as connectivity, distance, shape, and volume remains unclear. This limitation becomes particularly problematic in medical applications where topological understanding is critical for accurate diagnosis and treatment planning.

The need for specialized benchmarks becomes particularly evident when considering applications that require understanding of connectivity, distance relationships, and topological properties of spatial structures. Current approaches often fail to capture the nuanced requirements of spatial understanding, leading to systems that may perform well on standard benchmarks but fail in real-world scenarios requiring geometric reasoning.

Recent work in spatial logic and model checking has demonstrated the importance of formal specifications for spatial properties, (see \cite{CBLM21} and the citations therein). Spatial model checking is a novel image analysis technique, borrowed from the tradition of Formal Methods in Computer Science (see, \cite{baier2008principles}), that uses logic formulas to identify the pixels of an image that satisfy certain topological properties, or imaging features, such as connectedness, proximity, metric distance or texture similarity concerning a target region, and so on. The tool VoxLogicA, \cite{BCLM19}, \cite{BCM25}, provides a practical framework for spatial model checking in image analysis applications, \cite{CBLM21}, offering both theoretical foundations and practical tools for spatial reasoning tasks. The framework has been successfully applied to medical image analysis and spatial verification tasks, demonstrating its effectiveness in capturing complex spatial relationships, including brain tissue segmentation and hybrid AI approaches combining symbolic reasoning with machine learning techniques. However, the integration of such formal approaches with neural network evaluation remains largely unexplored. 

This paper proposes the design and a first work-in-progress evaluation of a comprehensive benchmark framework specifically designed to evaluate spatial reasoning capabilities in neural networks. We use as a prime example the state-of-the-art medical image segmentation network nnU-Net, \cite{IJKPM21}. 
The framework is designed to be easily expanded and currently provides two complementary task categories: maze connectivity analysis and spatial distance computation, plus their combination. The framework's multi-resolution approach enables systematic evaluation of scalability and generalization properties, while its automated pipeline ensures reproducibility and standardization across experiments.

\section{Related Work}
Despite the widespread use of overlap and distance metrics for segmentation evaluation (e.g., Dice coefficient, IoU, Hausdorff distance), these measures exhibit well-known limitations for assessing spatial structure understanding. Dice coefficient quantifies only volumetric overlap and remains insensitive to topological properties: two segmentations with identical Dice scores may differ dramatically in connectivity or shape characteristics, \cite{RTB24}. Similarly, Hausdorff distance captures worst-case boundary error but ignores global structural relationships and demonstrates high sensitivity to outliers, \cite{CBOZS22}. Recent comprehensive analyses have highlighted that common evaluation metrics disregard complex spatial relationships and fail to detect when segmentations break fundamental connectivity properties, \cite{RTB24}. Consequently, major structural errors such as missing branches or topological holes can remain undetected by standard benchmarking approaches, prompting calls for problem-aware metrics and additional topological evaluation beyond traditional overlap measures, \cite{RTB24}, \cite{CBOZS22}.

In medical imaging applications, the lack of explicit spatial reasoning leads to clinically relevant failures. Many anatomical targets including vessels, airways, neural fibers, and organ shapes exhibit tubular or tree-like structures where preserving connectivity is vital for accurate analysis. Studies have demonstrated that vessel segmentations achieving high Dice scores can be fundamentally different: one may miss critical small branches while another preserves the complete vascular network, with only the latter guaranteeing topology preservation, \cite{S21}. Even high-resolution 3D segmentation models achieving DSC $\approx$ 0.95 still exhibit decreased connectivity in fine vascular structures, \cite{YAJ24}, potentially invalidating downstream analyses such as blood flow modeling.

To address these limitations, recent work has incorporated explicit topology and geometry-aware components into neural network architectures. Notable strategies include topology-aware loss functions based on topological invariants such as Betti numbers from persistent homology, which force networks to respect global structural properties, \cite{CBOZS22}, \cite{S21}, \cite{PHADDED}. Graph-based connectivity networks represent another significant advancement, embedding explicit graph reasoning into segmentation pipelines. Hybrid CNN+GCN approaches combine standard convolutional networks (CNN) with graph convolutional networks (GCN) over anatomical skeleton points, learning global structural adjacency to improve connectivity preservation, particularly in complex vascular and retinal imaging applications, \cite{SLYL19}. Shape and distance priors have proven effective through the incorporation of anatomical knowledge and auxiliary geometric cues. Statistical shape models guide segmentation toward plausible contours, \cite{SLZGHZ24}, while distance transform supervision and boundary-aware losses capture proximity and boundary information that generic pixel-level supervision fails to capture. Topology refinement and post-processing approaches apply secondary models to correct topological errors, demonstrating reliable improvements in connectivity preservation across different base architectures, \cite{L24}.

Current benchmarking approaches in medical image segmentation predominantly rely on pixel-wise metrics that inadequately assess spatial reasoning capabilities. Recent advances in spatial reasoning evaluation have highlighted both the limitations of existing approaches and promising new directions \cite{THO22}, such as graph-based relational reasoning frameworks for spatial configuration recognition, message-passing neural architectures for geometric similarity tasks, and systematic diagnostic benchmarks that evaluate spatial understanding across varying object complexity levels, yet these methods remain fragmented and lack systematic integration into comprehensive evaluation frameworks. While topology-preserving losses, graph-based connectivity modules, and spatial shape priors demonstrate significant improvements in preserving anatomical structure, systematic evaluation frameworks for spatial reasoning remain limited. 
This gap motivates the development of comprehensive benchmarks that can systematically assess neural network understanding of fundamental spatial relationships across different scales and complexity levels, forming the foundation for the framework presented in this work.

\section{Problem Formulation}

Recent empirical studies have provided compelling evidence of systematic limitations in current neural architectures when confronted with spatial reasoning tasks, revealing fundamental gaps in geometric understanding, \cite{liu2023visual}.

The central research question addressed in this work concerns the reliability of neural networks in understanding morphological properties of images, specifically focusing on connectivity, distance relationships, shape analysis, and volumetric understanding. This systematic approach to spatial reasoning evaluation represents a departure from traditional benchmarking methods that rely on overlap metrics, providing instead a framework for testing specific spatial capabilities in isolation.

The problem can be formulated as follows: given a neural network trained on spatial reasoning tasks, how reliably can it identify and analyze basic morphological properties such as path connectivity in maze-like structures and distance relationships between spatial points? This question becomes critical in medical imaging applications, where topological understanding directly impacts diagnostic accuracy and treatment effectiveness, \cite{CBOZS22}, \cite{BCVMK23}. For instance, recent research has shown that networks enhanced with topological constraints—such as persistent homology or distance-based losses—demonstrate improved segmentation fidelity, preserving key features like anatomical connectivity and continuity in vascular or bronchial structures, \cite{MWWCH21}. Furthermore, architectures like nnU-Net, while highly performant, benefit from post-processing modules that ensure topological correctness without requiring model retraining saving considerable computation time and memory resources.

\subsection{Spatial Reasoning Challenges}
Neural networks face several fundamental challenges in spatial reasoning tasks. This work focuses on two of these: first, the discrete nature of digital images can introduce ambiguities in connectivity analysis, especially at low resolutions where pixel-level decisions critically affect topological properties; second, distance computation requires an understanding of metric properties that are not naturally represented in convolutional architectures. These challenges are directly addressed in the experiments presented, as described in detail in the following sections. Additionally, a third issue, common to both experiments, is scale invariance, which can hinder the preservation of spatial relationships across different resolution levels.

\subsubsection{Multi-Resolution Complexity}
The multi-resolution aspect of spatial reasoning introduces additional complexity layers. Spatial relationships that are clearly defined at high resolutions may become ambiguous or impossible to determine at lower resolutions. This scaling behavior is particularly important for understanding the fundamental resolution requirements for different types of spatial reasoning tasks.

\section{Method}
\begin{figure}
    \includegraphics[width=0.48\textwidth]{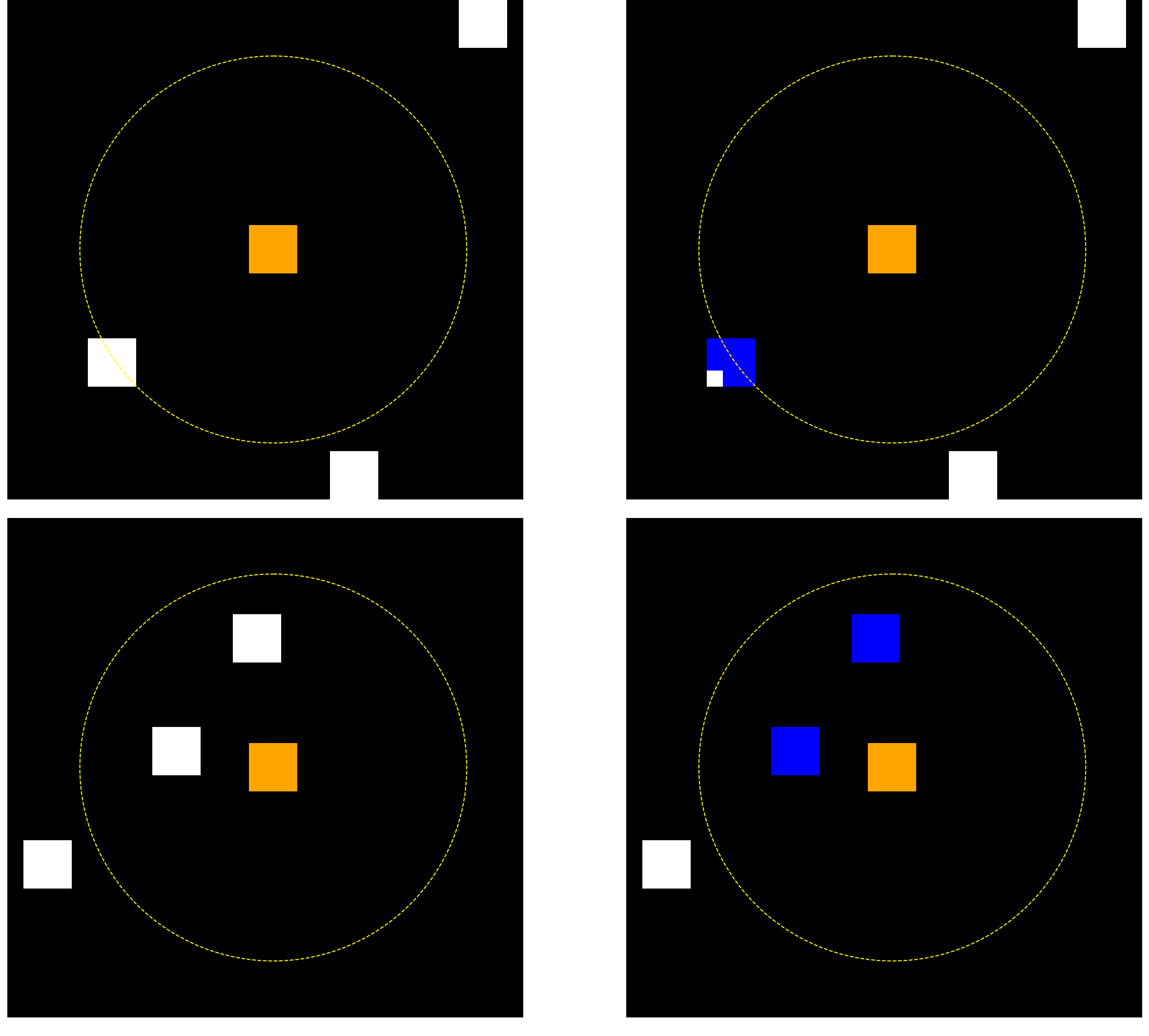}
    \caption{Examples from the Dots Distance dataset. The first column shows the two-channel input: orange squares mark the reference point (Channel 1) while white squares represent dot locations (Channel 0). The second column displays the corresponding ground truth segmentations (blue dots) identifying dots within the distance threshold (i.e., inside the yellow circle). The dataset thus enables supervised learning of spatial distance relationships. Two 32px resolution examples ("dots\_029" and "dots\_030") are shown.}
    \label{fig:dots}
\end{figure}

\begin{figure}
    \includegraphics[width=0.48\textwidth]{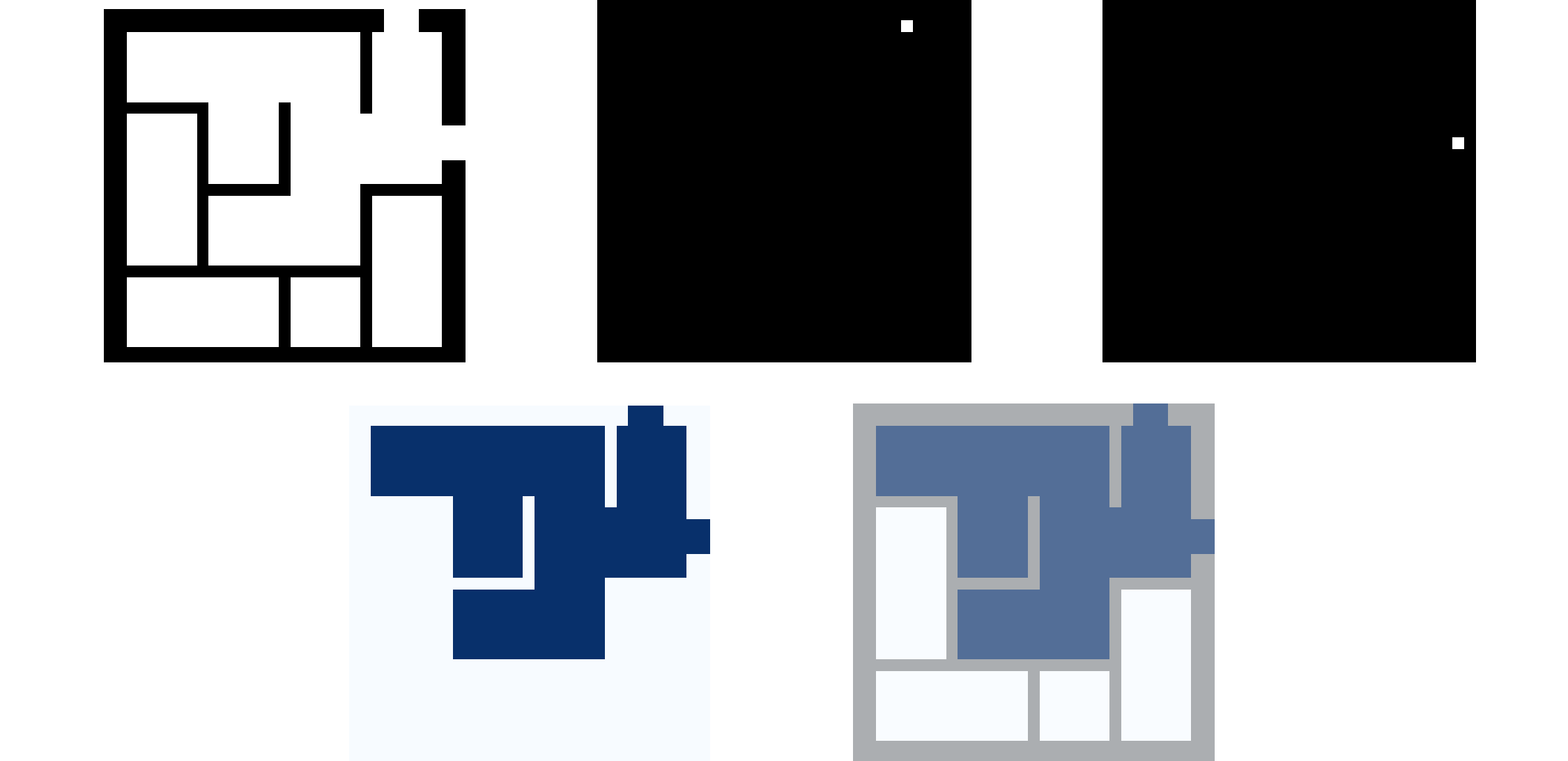}
    \caption{Three-channel encoding for maze navigation tasks. Training sample from the 4×4 maze dataset shown in the top row: spatial maze structure and entry/exit locations encoded in separate channels, and in the bottom row: corresponding ground truth path computed through a VoxLogicA spatial logic specification. The case "maze\_4x4\_048" at 32 px resolution is shown as an example. The three-channel input format (Channel 0: maze structure; Channel 1: entry point; Channel 2: exit point) enables nnU-Net to learn spatial navigation through supervised learning on logically-derived optimal paths.}
    \label{fig:maze}
\end{figure}

To address these challenges, we developed a comprehensive benchmark framework that systematically evaluates neural network performance on well-defined spatial reasoning tasks. 
Our key methodological innovation lies in combining formal spatial logic specifications with controlled synthetic dataset generation, enabling mathematically precise ground truth definition while maintaining complete control over spatial complexity.
The complete benchmark framework is designed to encompass multiple test categories evaluating various spatial characteristics including the shape and size of connected regions, connectivity patterns, distance relationships, spatial inclusion properties, region counting, and other fundamental geometric and topological features. The current implementation presents two representative task categories—maze connectivity analysis and spatial distance computation—as initial components of this broader evaluation system. The framework is built around three core components: automated dataset generation using VoxLogicA logical specifications (Figure \ref{fig:dots} and Figure \ref{fig:maze}), standardized training protocols, and comprehensive evaluation metrics that capture both pixel-level accuracy and spatial reasoning quality. In our first experimental setup, we have started applying our benchmark to nnU-Net to evaluate its performance in these dimensions, \cite{IJKPM21}, leveraging established methodologies for adaptive neural architectures and multi-resolution analysis, \cite{isensee2024nnu}, \cite{fan2019multiscale}.

\subsection{Multi-Resolution Approach}
The entire analysis was conducted through a systematic multi-resolution approach, evaluating neural network performance at 16px, 32px, and 64px for both task categories (maze connectivity and dots distance). The networks were trained using 50 images for each case (see details in Table \ref{tab:dots} and Table \ref{tab:maze}). This methodology enables the identification of how spatial reasoning capabilities evolve with varying input granularity and determination of minimum resolution requirements for reliable spatial reasoning. Our multi-scale analysis approach builds upon established techniques for hierarchical feature processing and multi-resolution computational modeling, providing a systematic framework for evaluating spatial understanding across different scales of detail.

\begin{figure}
    \includegraphics[width=0.48\textwidth]{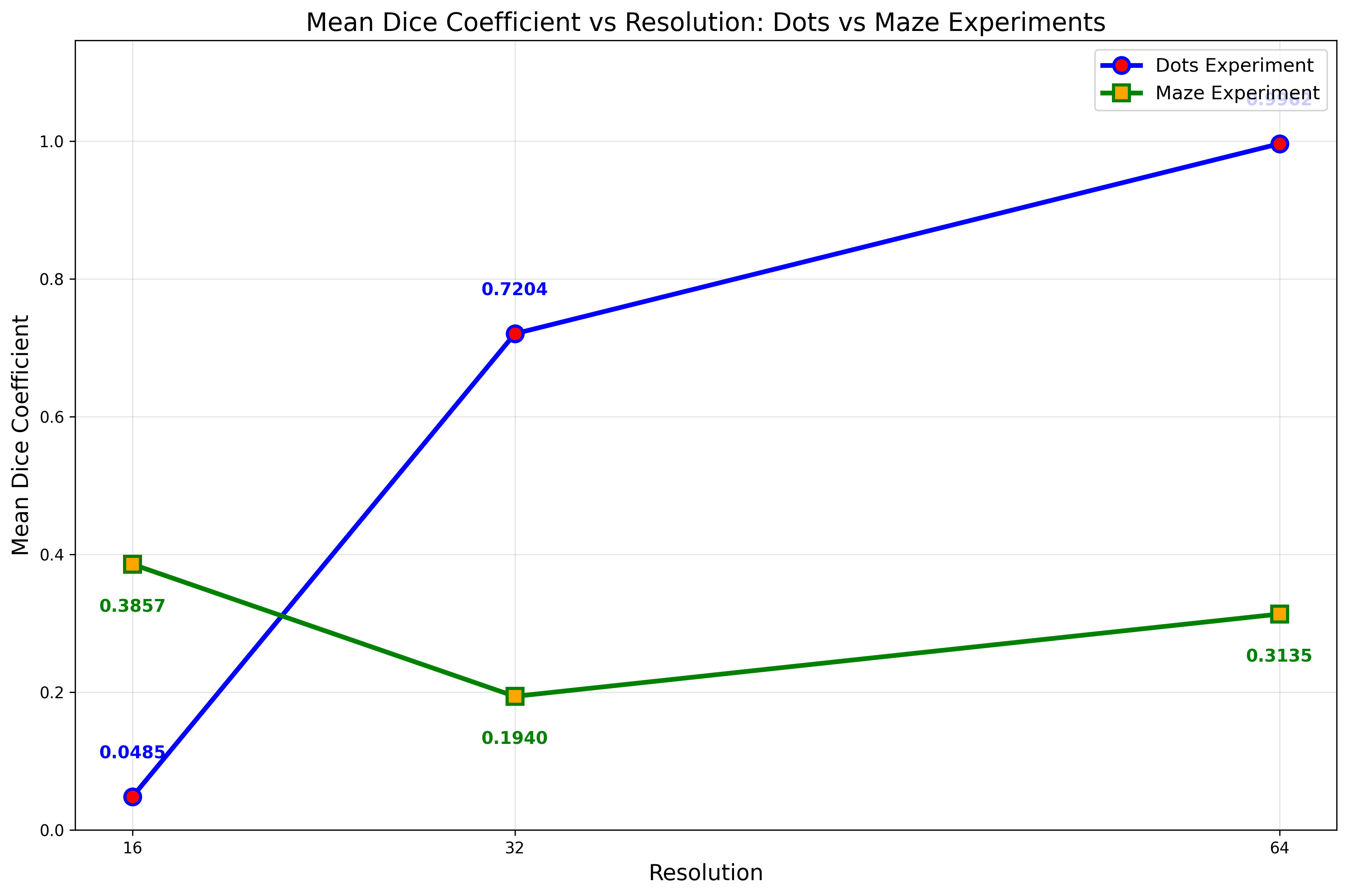}
    \caption{Performance comparison between dots and maze experiments as a function of image resolution. The mean Dice coefficient shows significant improvement for distance computation tasks (blue line) with increasing resolution, while connectivity analysis tasks (green line) maintain consistently low performance regardless of resolution, highlighting different challenges in spatial reasoning.}
    \label{fig:performance}
\end{figure}

As illustrated in Figure \ref{fig:performance}, the mean Dice coefficient progression reveals dramatically different behaviors between the two task types. For the dots experiments, we observe a significant performance improvement with increasing resolution: from near-zero values at 16px (Dice = 0.0485) to excellent performance at 64px (Dice = 0.9962), with an intermediate value at 32px (Dice = 0.7204). In contrast, the maze experiments show consistently low and relatively stable performance across all resolutions, with a slight decrease from 16px (Dice = 0.3857) to 32px (Dice = 0.1940) followed by modest recovery at 64px (Dice = 0.3135). These results \footnote{Note that the Dice scores should not be taken for granted, especially the positive results, as a more comprehensive assessment is required to guarantee bounds on the quality of segmentation under spatial constraints; our research so far has been focused on establishing a reproducible benchmarking system, and not yet on its full validation.}, although still preliminary, to the best of our knowledge reveal previously unreported behavior in neural network spatial reasoning: while distance computation tasks benefit significantly from increased resolution, connectivity analysis tasks remain problematic regardless of input granularity, suggesting deeper limitations in neural networks' topological understanding capabilities.

\subsection{Implementation Methodology}
The proposed methodology represents a novel approach to spatial reasoning evaluation by creating controlled synthetic datasets where ground truth spatial relationships are unambiguously defined through logical specifications rather than manual annotation or heuristic generation.

This approach enables systematic evaluation of neural network performance while maintaining complete control over problem complexity and spatial properties. The current framework encompasses two complementary evaluation domains that assess different dimensions of spatial understanding: connectivity analysis through maze navigation tasks and proximity reasoning via distance-based classification.

Our approach leverages VoxLogicA, \cite{BCLM19}, \cite{BCM25}, a spatial logic model checker, featuring a declarative specification language, to create ground truth labels that precisely capture the spatial relationships under investigation. The use of formal spatial logic ensures that our benchmark tasks have mathematically precise definitions, eliminating ambiguity in ground truth generation. For the maze experiments, the system generates 4x4 grid structures with configurable connectivity parameters, where connectivity relationships are formally specified using VoxLogicA's spatial operators. The dots experiments focus on distance-based segmentation, where spatial proximity relationships are defined through VoxLogicA's distance predicates.

\subsubsection{Dataset generation}
Our experimental framework generates two distinct types of synthetic datasets as examples of possible tasks to evaluate neural network segmentation capabilities on logical reasoning. The use of synthetic data for benchmark construction has gained significant validation in recent machine learning research, with studies demonstrating that well-designed synthetic datasets can provide reliable evaluation frameworks for complex reasoning tasks.

Dots distance datasets are composed of synthetically generated images containing multiple dots with varying radii and spatial positions. The segmentation target is defined as the set of regions within a specified distance from reference dots, with ground truth masks computed using VoxLogicA’s formal spatial logic predicates. Each sample consists of a two-channel input image: the first channel encodes the reference dot, while the second channel encodes the set of white dots. The corresponding ground truth is a single-channel mask highlighting all pixels within the computed distance threshold from the reference dot. This approach allows for precise control over spatial relationships and enables systematic variation of parameters such as dot radius, number of dots, and distance thresholds, facilitating the exploration of different levels of task complexity.

Maze datasets are procedurally generated as 4×4 grid mazes with configurable connectivity parameters (e.g., connection probability p=0.7 for partial connectivity). For each maze instance, the ground truth solution path is computed using VoxLogicA’s spatial operators, ensuring a mathematically rigorous definition of connectivity. The input to the network is a three-channel image encoding the maze walls, entrance, and exit, while the ground truth is a single-channel mask representing the solution path from entrance to exit. The procedural generation process allows for the creation of a diverse set of mazes with varying topological properties, supporting robust evaluation of neural network performance on connectivity reasoning tasks.

All datasets are generated across multiple resolutions (16px, 32px, 64px) with configurable parameters: dataset size, dots-specific parameters (radius scaling, number scaling, distance thresholds), and maze-specific parameters (connectivity type and connection probability). The dataset generation process is fully automated and integrated within the benchmark pipeline, ensuring consistency, reproducibility, and scalability. This setup enables systematic investigation of how neural network performance varies with spatial granularity and task complexity and provides a flexible foundation for future extensions to additional spatial reasoning tasks.

\subsubsection{Pipeline automation and reproducibility}
All stages of the experimental workflow—including dataset generation, model training, inference, and quantitative evaluation—are seamlessly integrated and orchestrated through a unified, modular Python pipeline. This pipeline is designed to maximize reproducibility and flexibility, enabling researchers to execute the entire benchmark or individual components via a single command-line interface.
The pipeline supports extensive parameterization, allowing users to specify experiment type (maze or dots), spatial resolutions, dataset sizes, neural network training parameters (such as number of folds for cross-validation), and computational resources (CPU or GPU). All configuration options are centrally managed, ensuring consistency across experiments and facilitating systematic exploration of different settings.\\
Model training is performed in batch mode across all relevant datasets, with automated detection of available data and robust error handling. Inference and evaluation steps are likewise fully automated, with results systematically organized in structured directories to support downstream analysis and direct comparison between experimental conditions.
The modular design of the pipeline, with clearly separated functions for each stage, enables straightforward adaptation to new experimental scenarios or the addition of further spatial reasoning tasks. This approach not only ensures the reproducibility of results but also provides a scalable and extensible foundation for future research in neural network-based spatial reasoning.\\
The entire pipeline, including all scripts and configuration files, is released as open-source software to promote transparency, reproducibility, and community-driven development, \footnote{The multi-experiment benchmarking pipeline is available at \url  {https://github.com/Manuelaimbriani12/nnUNet_benchmark}.}.

\subsubsection{Evaluation and Representative Results}
Performance evaluation is conducted using established segmentation metrics including Dice coefficient and Intersection over Union (IoU), supplemented by specialized visualization tools that enable detailed analysis of prediction quality and failure modes. As demonstrated in our experimental results, even state-of-the-art neural networks, validated using traditional performance metrics such as the Dice score, show significant limitations in basic spatial reasoning tasks (Table \ref{tab:dots} and Table \ref{tab:maze}).

\begin{table}
\caption{Preliminary dots dataset results across pixel resolutions (16px, 32px, 64px). Values represent mean $\pm$ standard deviation for IoU and Dice coefficients (N=50 per experiment). Initial findings suggest performance improvement with increased resolution, though results require validation with additional metrics.}
\label{tab:dots}
\begin{ruledtabular}
\begin{tabular}{cccc}
\textbf{Resolution} & \textbf{N.cases} & \textbf{IoU} & \textbf{Dice} \\
\hline
16px & 50 & 0.0264 $\pm$ 0.0428 & 0.0485 $\pm$ 0.0723 \\
32px & 50 & 0.7069 $\pm$ 0.4333 & 0.7204 $\pm$ 0.4352 \\
64px & 50 & 0.9924 $\pm$ 0.0115 & 0.9962 $\pm$ 0.0059 \\
\end{tabular}
\end{ruledtabular}
\end{table}

\begin{table}
\caption{Preliminary maze dataset results across pixel resolutions (16px, 32px, 64px). Values represent mean $\pm$ standard deviation for IoU and Dice coefficients (N=50 per experiment). Initial results indicate best performance at lowest resolution, though high variability suggests need for more appropriate evaluation metrics.}
\label{tab:maze}
\begin{ruledtabular}
\begin{tabular}{cccc}
\textbf{Resolution} & \textbf{N.cases} & \textbf{IoU} & \textbf{Dice} \\
\hline
16px & 50 & 0.3587 $\pm$ 0.4334 & 0.3857 $\pm$ 0.4574 \\
32px & 50 & 0.1771 $\pm$ 0.3482 & 0.1940 $\pm$ 0.3706 \\
64px & 50 & 0.2951 $\pm$ 0.4260 & 0.3135 $\pm$ 0.4432 \\
\end{tabular}
\end{ruledtabular}
\end{table}

\begin{figure}
    \includegraphics[width=0.48\textwidth]{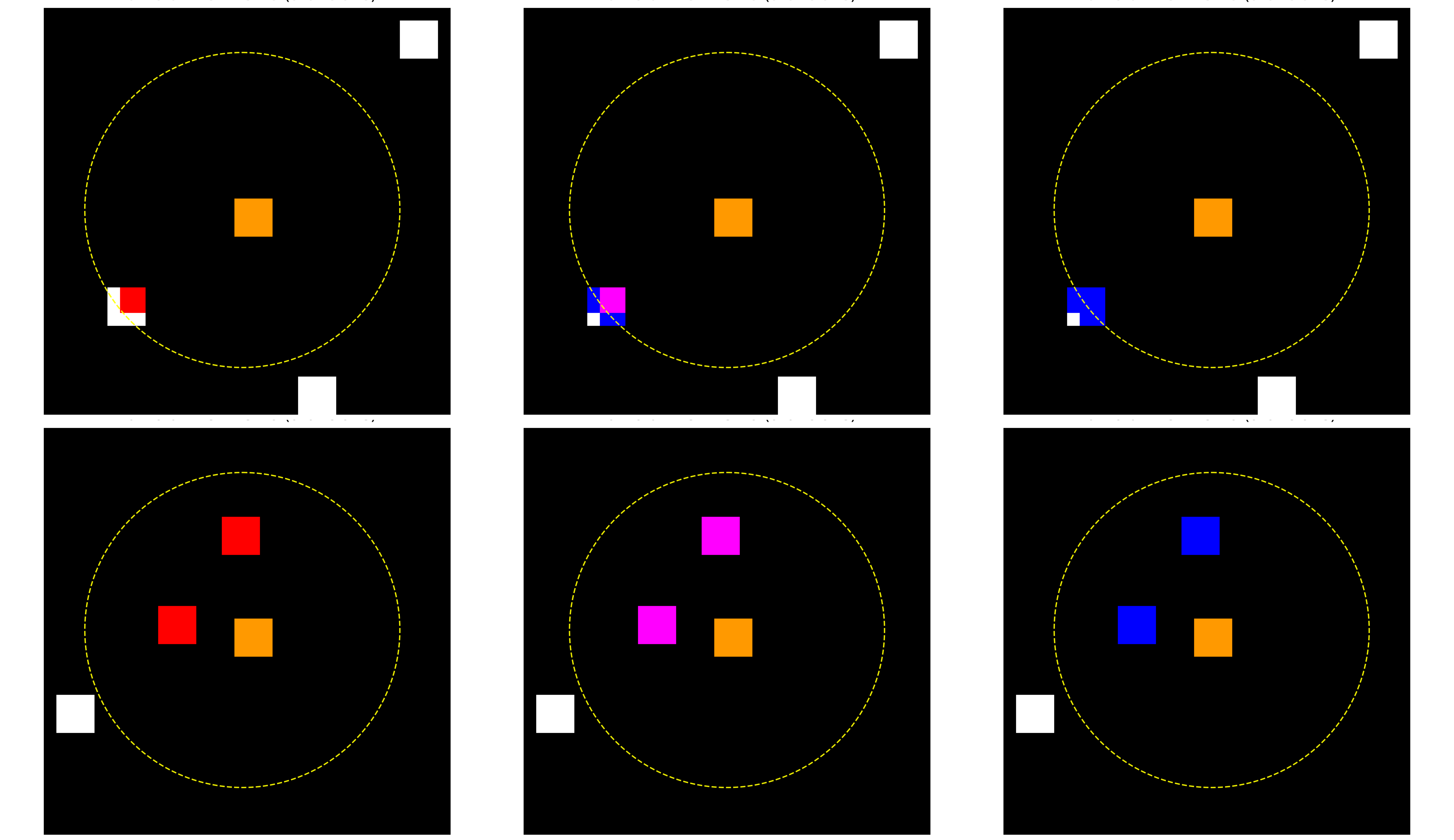}
    \caption{Visual comparison of neural network segmentation performance on two representative cases at 32px from the dots experiment (dots\_029 and dots\_030). Each row displays three visualization types: prediction (left), prediction overlaid with ground truth label (center), and ground truth label (right). Top row shows results for case dots\_029 (IoU: 0.50, Dice: 0.67), bottom row shows results for case dots\_030 (IoU: 1.00, Dice: 1.00). The overlaid visualization in the center column facilitates direct comparison between predicted and actual segmentation boundaries, highlighting areas of agreement (overlap) and discrepancy between the model output and ground truth annotations. The quantitative metrics demonstrate varying performance levels, with dots\_030 achieving perfect segmentation while dots\_029 shows moderate accuracy with 50\% overlap.}
    \label{fig:dots_results}
\end{figure}

\begin{figure}
    \includegraphics[width=0.48\textwidth]{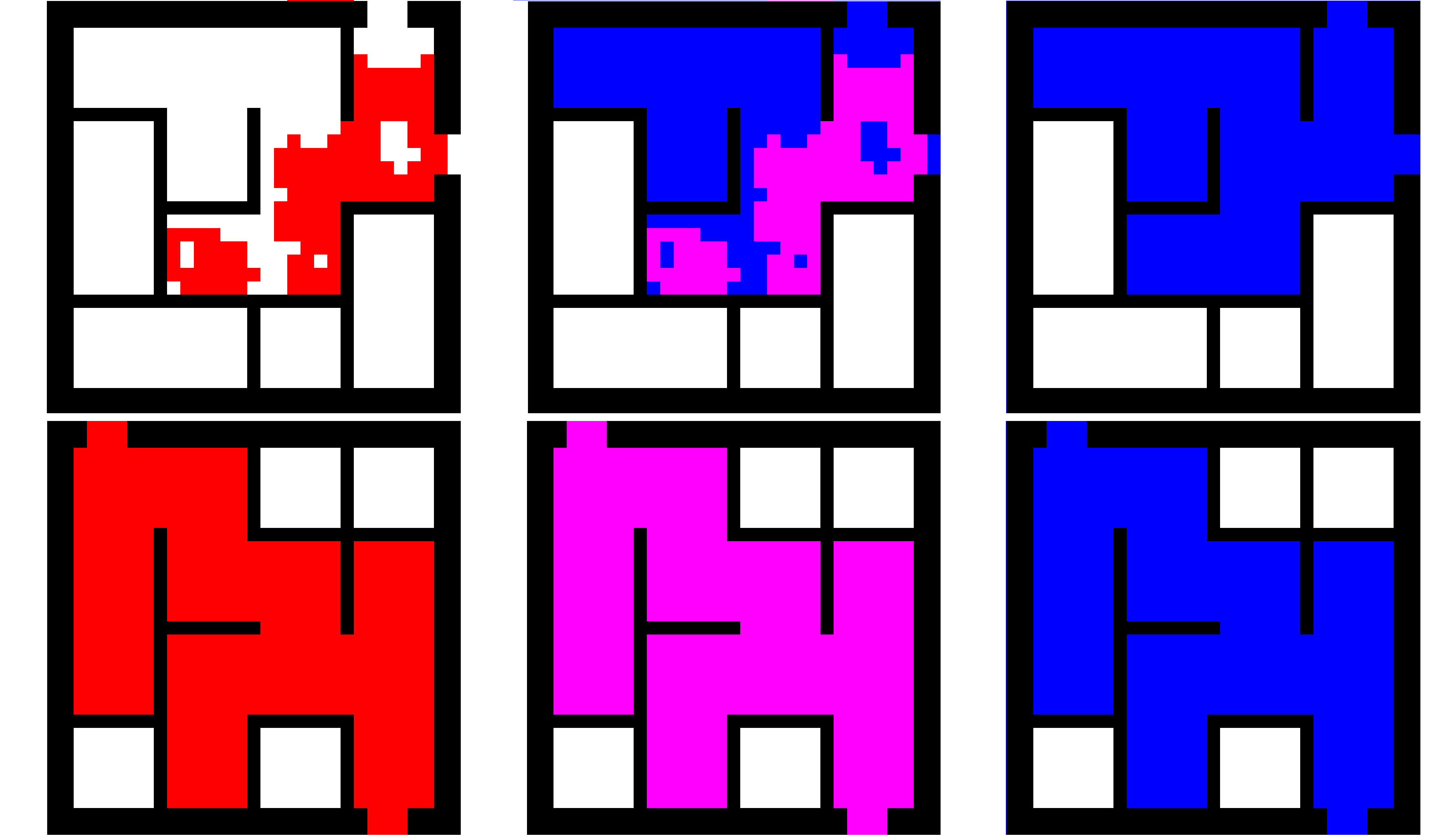}
    \caption{Comparative analysis of neural network segmentation performance on two selected cases from the 4x4 maze experiment at 32px (maze\_4x4\_048 and maze\_4x4\_034). The visualization follows the same three-column format: prediction (left), prediction with ground truth overlay (center), and ground truth label (right). Top row presents results for case maze\_4x4\_048 (IoU: 0.32, Dice: 0.49), while bottom row shows case maze\_4x4\_034 (IoU: 1.00, Dice: 1.00). The center column overlay enables quantitative visual assessment of segmentation accuracy by clearly delineating regions where the predicted segmentation aligns with or deviates from the reference standard. The metrics reveal contrasting performance between cases, with maze\_4x4\_034 achieving perfect segmentation accuracy while maze\_4x4\_048 demonstrates partial overlap with approximately one-third intersection coverage.}
    \label{fig:maze_results}
\end{figure}

The following representative examples were evaluated at 32px resolution. Figure \ref{fig:dots_results} illustrates representative results from the dots experiment, showing both failure and success cases: dots\_029 demonstrates moderate performance with partial spatial understanding (IoU: 0.50, Dice: 0.67), while dots\_030 achieves perfect segmentation accuracy (IoU: 1.00, Dice: 1.00). Figure \ref{fig:maze_results} presents results from the maze connectivity analysis, revealing more pronounced performance variability: maze\_4x4\_048 shows substantial difficulty in identifying connected components with limited spatial reasoning capability (IoU: 0.32, Dice: 0.49), whereas maze\_4x4\_034 demonstrates complete topological understanding (IoU: 1.00, Dice: 1.00). These contrasting initial results within each task category highlight the inconsistent nature of neural network spatial reasoning performance, where seemingly similar spatial configurations can yield dramatically different outcomes, underscoring the fundamental challenges in achieving reliable spatial understanding across diverse scenarios.

\section{Conclusion}
The preliminary experimental results of this ongoing study clearly demonstrate that the spatial reasoning problem is not merely a theoretical concern but a practical limitation that manifests itself in real-world applications. 
Our systematic evaluation approach, combining formal spatial logic with multi-resolution testing, has revealed fundamental limitations in neural network spatial understanding that remain hidden by traditional evaluation metrics.
Neural networks trained on our benchmark datasets show systematic failures in basic connectivity and distance reasoning tasks, even when provided with relatively simple synthetic data. These initial findings may have significant implications for clinical applications where topological understanding is critical for accurate analysis.

The benchmark framework we are designing will provide a standardized platform for evaluating spatial reasoning limitations and serves as a foundation for developing improved approaches. Our preliminary results suggest that pure neural network approaches may be insufficient for reliable spatial reasoning, pointing toward the need for hybrid methods that combine neural networks with symbolic reasoning capabilities. This aligns with recent trends in AI research that emphasize the integration of connectionist and symbolic approaches for robust intelligent systems, \cite{THO22}.

Recent work has further demonstrated the promising potential of hybrid approaches in medical imaging applications, where the combination of neural network pattern recognition capabilities with formal spatial reasoning methods can address the fundamental limitations identified in our evaluation, \cite{BBCLM24}.

While these initial results are promising, several aspects of this work-in-progress require further development and refinement. The current evaluation is limited to relatively simple synthetic datasets with basic spatial configurations, and validation on more complex real-world scenarios is necessary to fully understand neural network spatial reasoning capabilities. Additionally, the framework currently focuses on two specific types of spatial reasoning tasks and would benefit from expansion to include additional categories such as shape manipulation, volumetric understanding, and multi-scale spatial relationships.

A significant component of our future work will focus on investigating and developing novel performance metrics that properly account for the geometric and morphological properties addressed by our benchmark framework. Current evaluation metrics such as Dice coefficient and IoU, while valuable for basic overlap assessment, are fundamentally inadequate for capturing spatial reasoning quality, topological correctness, and morphological understanding. This limitation necessitates extensive literature review and potential development of specialized metrics that can quantify connectivity preservation, shape fidelity, distance relationship accuracy, and other spatial reasoning capabilities that our benchmark framework is designed to assess. Such metrics are essential for meaningful evaluation of neural network spatial reasoning performance and will be crucial for guiding the development of improved architectures and training methodologies.

The key innovation of our approach lies in the systematic combination of formal spatial logic specifications with neural network evaluation, enabling precise identification of spatial reasoning failures that traditional metrics cannot detect. Our multi-resolution analysis reveals interesting task-dependent scaling behaviors that warrant further investigation.

This ongoing research will be extended in several directions. We are actively developing extensions to our benchmark framework, with ongoing work on shape-based reasoning tasks. However, scaling our formal specification approach to these more complex spatial domains requires careful design to preserve the mathematical precision that enables our current systematic evaluation.
Future work will focus on expanding the benchmark to include additional spatial reasoning tasks and developing hybrid methods that address the identified limitations. The framework's modular design enables straightforward extension to new task categories and evaluation metrics, supporting continued advancement in spatial reasoning research. We are currently investigating the integration of symbolic reasoning methods with neural network architectures as a promising direction for achieving reliable spatial understanding in clinical and other critical applications, \cite{BCM25}, \cite{BBCLM24}. The open-source implementation and comprehensive documentation enable widespread adoption and facilitate comparative studies across different architectural approaches. By providing a rigorous evaluation framework, this work-in-progress contributes to the development of more reliable AI systems capable of robust spatial reasoning in real-world applications, though significant research and development efforts remain to achieve this goal.

\begin{acknowledgments}
Research partially supported by Bilateral project between CNR (Italy) and SRNSFG (Georgia) "Model Checking for Polyhedral Logic" (\#CNR-22-010); European Union - Next GenerationEU - National Recovery and Resilience Plan (NRRP), Investment 1.5 Ecosystems of Innovation, Project "Tuscany Health Ecosystem" (THE), CUP: B83C22003930001; European Union - Next-GenerationEU - National Recovery and Resilience Plan (NRRP) – MISSION 4 COMPONENT 2, INVESTMENT N. 1.1, CALL PRIN 2022 D.D. 104 02-02-2022 – (Stendhal) CUP N. B53D23012850006; MUR project PRIN 2020TL3X8X "T-LADIES"; CNR project "Formal Methods in Software Engineering 2.0", CUP B53C24000720005; Shota Rustaveli National Science Foundation of Georgia grant.
\end{acknowledgments}

\bibliography{biblio2}

\end{document}